\pdfoutput=1

\documentclass[11pt]{article}

\usepackage[final]{acl}

\usepackage{times}
\usepackage{latexsym}

\usepackage[T1]{fontenc}

\usepackage[utf8]{inputenc}

\usepackage{microtype}

\usepackage{inconsolata}
\usepackage{graphicx}
\usepackage{amsmath}
\usepackage{multirow}

\usepackage{algorithm}
\usepackage{algorithmic}

\title{An Investigation Into Explainable Audio Hate Speech Detection}

\author{Jinmyeong An $^*$$^1$, Wonjun Lee $^*$$^2$, Yejin Jeon $^{1}$, \\
        \textbf{Jungseul Ok $^{1,2}$, Yunsu Kim $^{3}$ \and Gary Geunbae Lee $^{1,2}$}\\
        $^1$ Graduate School of Artificial Intelligence, POSTECH, Republic of Korea \\
	  $^2$ Department of Computer Science and Engineering, POSTECH, Republic of Korea \\
        $^3$ aiXplain Inc., Los Gatos, CA, USA \\
        \{jinmyeong, lee1jun, jeonyj0612, jungseul.ok, gblee\}@postech.ac.kr,\\ yunsu.kim@aixplain.com}
      
\begin{document}
\maketitle
\def\thefootnote{*}\footnotetext{Equally contributed}\def\thefootnote{\arabic{footnote}}
\begin{abstract}
    Research on hate speech has predominantly revolved around detection and interpretation from textual inputs, leaving verbal content largely unexplored. While there has been limited exploration into hate speech detection within verbal acoustic speech inputs, the aspect of interpretability has been overlooked. Therefore, we introduce a new task of explainable audio hate speech detection. Specifically, we aim to identify the precise time intervals, referred to as audio frame-level rationales, which serve as evidence for hate speech classification. Towards this end, we propose two different approaches: cascading and End-to-End (E2E). The cascading approach initially converts audio to transcripts, identifies hate speech within these transcripts, and subsequently locates the corresponding audio time frames. Conversely, the E2E approach processes audio utterances directly, which allows it to pinpoint hate speech within specific time frames. Additionally, due to the lack of explainable audio hate speech datasets that include audio frame-level rationales, we curated a synthetic audio dataset to train our models. We further validated these models on actual human speech utterances and found that the E2E approach outperforms the cascading method in terms of the audio frame Intersection over Union (IoU) metric. Furthermore, we observed that including frame-level rationales significantly enhances hate speech detection accuracy for the E2E approach.
    
  \textbf{Disclaimer} The reader may encounter content of an offensive or hateful nature. However, given the nature of the work, this cannot be avoided.
\end{abstract}

\section{Introduction}

Online platforms such as YouTube, Dailymotion, and TikTok have undoubtedly experienced a notable surge in popularity over the years. While this has led to an increased dependence on audio as a primary mode of communication, this phenomenon has also brought the issue of hate speech in audio content to the forefront. YouTube, for instance, has consistently been proactive in removing hateful content since its inception, aligning with its hate speech policy\footnote{https://blog.youtube/inside-youtube/the-four-rs-of-responsibility-remove/}. Nevertheless, it is worth noting that out of a total of 10,501,072 channels removed from the YouTube platform within the period of July to September 2023, 26,130 channels were specifically taken down due to their association with hate speech\footnote{https://transparencyreport.google.com/youtube-policy/removals?hl=en}. These statistics underscore the unequivocal importance and the imperative need for the development of effective methodologies to precisely identify hate speech within verbal expressions.

An important point to note, however, is that most hate speech datasets are exclusively text-based. Consequently, research endeavors pertaining to hate speech detection \cite{Qian-text, Park-text} as well as investigations into hate speech explainability \cite{HateXplain} are confined to textual inputs. In other words, despite the explosive increase of hate speech on audio-based online social platforms, there is a notable absence of research that addresses hate speech in verbal data. A few studies related to auditory hate speech detection have been proposed. For example, \citet{Audio1, Audio2-emotion} curated respective audio-visual multi-modal datasets. Yet, to the best of our knowledge, no research addresses \textit{explainable} hate speech detection in the audio domain, that is, the understanding of the rationale behind the model's decisions.

Therefore, we first introduce the new task of explainable audio hate speech detection, which encompasses two sub-tasks: audio hate speech classification (AHS-CLS) and audio hate speech frame detection (AHS-FD). The former involves determining whether an audio utterance is hate speech, while the latter identifies the specific time frames containing hate speech. In addition, since there is a lack of interpretable audio hate speech datasets that include audio frame-level rationales, we curated a dataset called AudioHateXplain, which annotates which part of human and synthetic audio recordings pertains to hate speech. Moreover, we propose cascading and E2E models, which are able to elucidate the underlying reasons for classifying speech as hate or not by identifying relevant audio rationales. The cascading approach first transforms audio into text transcripts, detects hate speech within these transcripts, and then matches the detected hate speech to the corresponding audio time frames. In contrast, the E2E approach directly processes the audio input, enabling it to identify the specific time frames containing hate speech accurately. We validated these models on actual human speech utterances and found that the E2E approach outperforms the cascading method regarding the audio frame Intersection over Union (IoU) metric. This superiority is attributed to the bottlenecks arising from conversion between audio and text, such as Automatic Speech Recognition (ASR) errors, and disarrangement between word tokens and time frames. Furthermore, we observed that including frame-level rationales significantly enhances hate speech detection accuracy for the E2E approach.

\section{Related Work}


\subsection{Hate Speech Detection}
Over the years, there have been considerable efforts towards text-based hate speech research, a domain that has gone through various and separate nomenclatures such as cyber hate, offensive, and online abusive \citet{abusive,abusivehatespeech} language detection. In this paper, we define these terms collectively as hate speech. In the initial stages of hate speech detection research, \citet{Smokey-1997} predominantly employed feature-based rules. Similarly, \citet{FlameDetection2} incorporated a set of rules to extract semantic information. More recently, in response to the escalating prevalence of online hate speech, there has been a concerted effort to curate private or publicly accessible hate speech datasets \citet{dataset1, dataset2}. However, training hate speech detection models on such datasets, which feature binary-level hate speech annotation, lacks interpretability. As a result, it becomes difficult to comprehend the logic behind model decisions. In light of this, \citet{HateXplain} curated a dataset with word-level annotations (rationales), which deviates from conventional datasets focused solely on increasing sentence-level model classification performance.

It is even more important to note that the predominant focus has been on text-based classification. In other words, more datasets and research for verbal hate speech detection and explanations must be needed. To address this, \citet{Audio1, Audio2-emotion} curated audio-visual multi-modal datasets, while \citet{Audio1} amassed short-form Filipino videos and compared different classification methods, including Support Vector Machine, logistic regression, and Random Forest. Similarly, \citet{Audio2-emotion} collected videos from Twitter and YouTube, then implemented a multi-task learning model to better identify hate speech by combining text, visual, and acoustic information.

Yet, to the best of our knowledge, no research addresses \textit{explainable} hate speech detection in the audio domain, that is, understanding the rationale behind the model's decisions. Our research endeavors extend beyond conventional text-based approaches by expanding into the audio domain. Moreover, we address model explainability in addition to audio hate speech detection.

\subsection{Audio Classification \& Frame Detection}

Classifying audio clips into specific categories, such as speech commands \cite{speech_command}, urban sound events \cite{ESC}, and the emotional content of speakers \cite{IEMOCAP}, has been extensively researched. In addition to classifying entire audio clips, there's been exploration into classifying audio frames at specific time intervals (e.g., every 10 milliseconds), as seen in speaker diarization studies \cite{CALLHOME, fujita2019end}. In this study, we combine both approaches to classify entire audio clips as containing hate speech or normal speech, while also pinpointing the exact segments within the audio where hate speech occurs, using a 10-millisecond time grid.

\begin{table}[t]
\centering
\begin{tabular}{ccc}
\hline
                     & \textbf{Samples}  & \textbf{Avg. Length (sec.)} \\ \hline
\textbf{Train}       & 14,183            & 6.62                        \\
\textbf{Dev.}        & 1,771             & 6.60                        \\ 
\textbf{Test-Synth.} & 300               & 8.97                        \\
\textbf{Test-Human}  & 300               & 10.52                       \\ \hline
\end{tabular}
\caption{Summary of the AudioHateXplain dataset, including audio duration. `Test-Synth' refers to the spoken audio generated by a text-to-speech (TTS) model, while `Test-Human' refers to audio recorded by human speakers.}
\label{tab:audiohatexplain}
\end{table}

\section{Dataset Generation}
Given the absence of existing explainable audio hate speech datasets, we created a synthetic dataset using a text-to-speech (TTS) model. This section details the methods used to convert text transcripts into spoken utterances and generate audio rationales for explaining audio hate speech.

To delineate the process of audio rationale generation, it is imperative first to comprehend the foundational structure of the original text-based HateXplain \cite{HateXplain} dataset. This text-based HateXplain dataset, represented as $\mathcal{D} = \{(x^{(1)}, W^{(1)}, y^{(1)}), \ldots, (x^{(L)}, W^{(L)}, y^{(L)})\}$, comprises $L$ samples. Each sample consists of a textual sentence $x$ paired with its corresponding binary class label $y \in \{0, 1\}$, denoting whether the sentence qualifies as hate speech (1) or as normal discourse (0). Moreover, each textual sentence $x$ is supplemented by a set of word-level annotations $W$, which is defined as $W = \{(w^{(1)}, \delta^{(1)}_w), \ldots, (w^{(N)}, \delta^{(N)}_w)\}$. Here, $N$ signifies the position of a word $w$ within the sentence, with each word linked to its word-level rationale $\delta_w \in \{0,1\}$. Specifically, a word $w$ is assigned a rationale of 1 if it contributes to the classification of the sentence as hate speech, and 0 otherwise. These word-level rationales serve as discernible evidence aiding in identifying and classifying hate speech within textual content.

\noindent\textbf{Text-to-Speech} From the above-mentioned text dataset, we convert each text-based transcript $x$ into audio samples of sample rate 22050 Hz. This conversion is achieved by leveraging the non-autoregressive FastSpeech \cite{FastSpeech2} TTS model in conjunction with the HiFi-GAN \cite{HifiGAN} vocoder. To ensure the coherence of audio samples, we expand abbreviations, remove emojis, and exclude sentences in languages other than English, as well as those that contain semantically vacuous words like the placeholder \textit{``\textless user\textgreater''}. 

\noindent\textbf{Rationale Labeling} Each TTS-generated audio sample $a$ is then paired with its binary classification label $y\in \{0, 1\}$ to identify whether it is hate speech (1) or normal audio (0). Moreover, it is imperative to provide acoustic rationales to facilitate the explainability of hate speech detection in the audio domain. To accomplish this, we employ a pretrained Montreal Forced Aligner \cite{MFA} to identify the timestamps, i.e., the starting and ending times in milliseconds of each spoken word within a given sentence. Subsequently, we divide audio samples into $M$ 10ms-long audio frames $f$, and each audio frame $f$ is annotated with its audio frame-level rationale $\delta_f \in \{0,1\}$. It is possible to annotate each audio frame with rationales as text-based word-level rationale $\delta_w$ are previously provided. Our AudioHateXplain dataset can thus be represented as $\mathcal{D} = \{(a^{(1)}, F^{(1)}, y^{(1)}), \ldots, (a^{(L)}, F^{(L)}, y^{(L)})\}$ of $L$ TTS-generated audio samples. In addition, the set of frame-level audio annotations $F$ can be reorganized as 
\begin{eqnarray}
F = \{(f^{(1)}, \delta^{(1)}_f), \ldots, (f^{(M)}, \delta^{(M)}_f)\}
\end{eqnarray}

\begin{figure*}[t]
    \centering
    \includegraphics[width=0.91\linewidth]{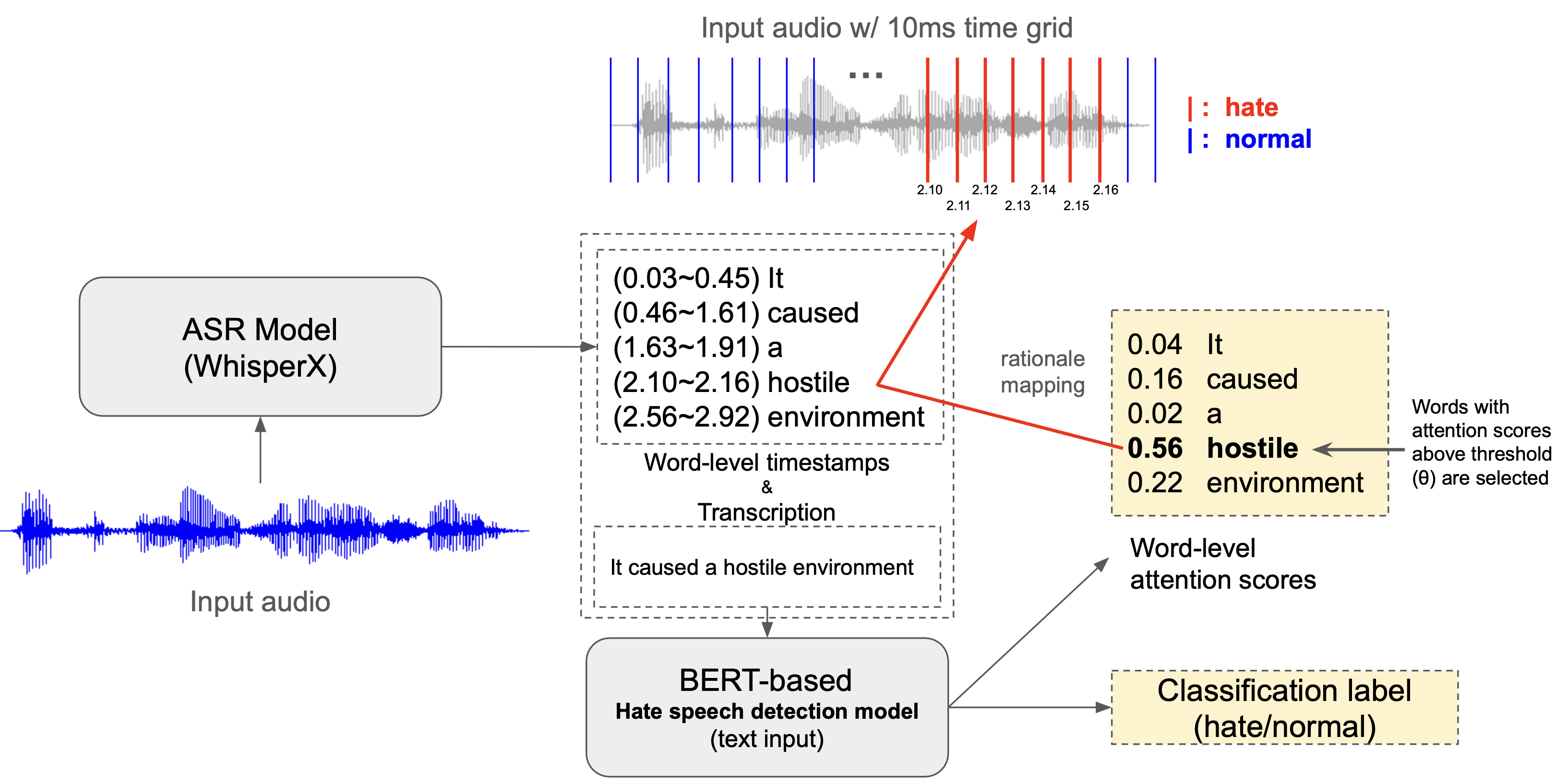}
    \caption{Overview of the cascaded method. Boxes highlighted in yellow indicate model outputs.}
    \label{fig:cas}
\end{figure*}

\noindent\textbf{Human Recordings} In addition to the synthetic audio dataset, we curated a separate collection of authentic human recordings for evaluation purpose. The transcript used for these recordings were derived from the original text-based HateXplain dataset, but underwent a two-step post-processing procedure. Specifically, among the 1,779 original test samples, we initially use ChatGPT (Appendix \ref{sec:PROMPTINGDETAILS}) to select texts that were suitable for spoken format. This process enabled us to sample 695 spoken-form texts from the HateXplain test set. These texts were then manually filtered to ensure that the final utterances adhered to syntactic and lexical choices appropriate for spoken language \cite{ong2002orality, Biber1986SpokenAW}. Ultimately, 300 utterances were selected for the test set. We synthesized these samples using TTS models and also recorded them with human participants.

The participant group consisted of 10 individuals, 6 males and 4 female speakers. Each participant read an average of 30 utterances, including hate speech and normal texts. Recordings were conducted in silent environments. For ethical considerations, all participants were fully informed about the nature of the transcripts, which included hateful language. Moreover, all recordings were conducted with the explicit consent of the volunteers for research purposes only.

The statistics of the AudioHateXplain is provided in Table \ref{tab:audiohatexplain}.

\section{Methodology}

\begin{figure*}[t]
    \centering
    \includegraphics[width=0.91\linewidth]{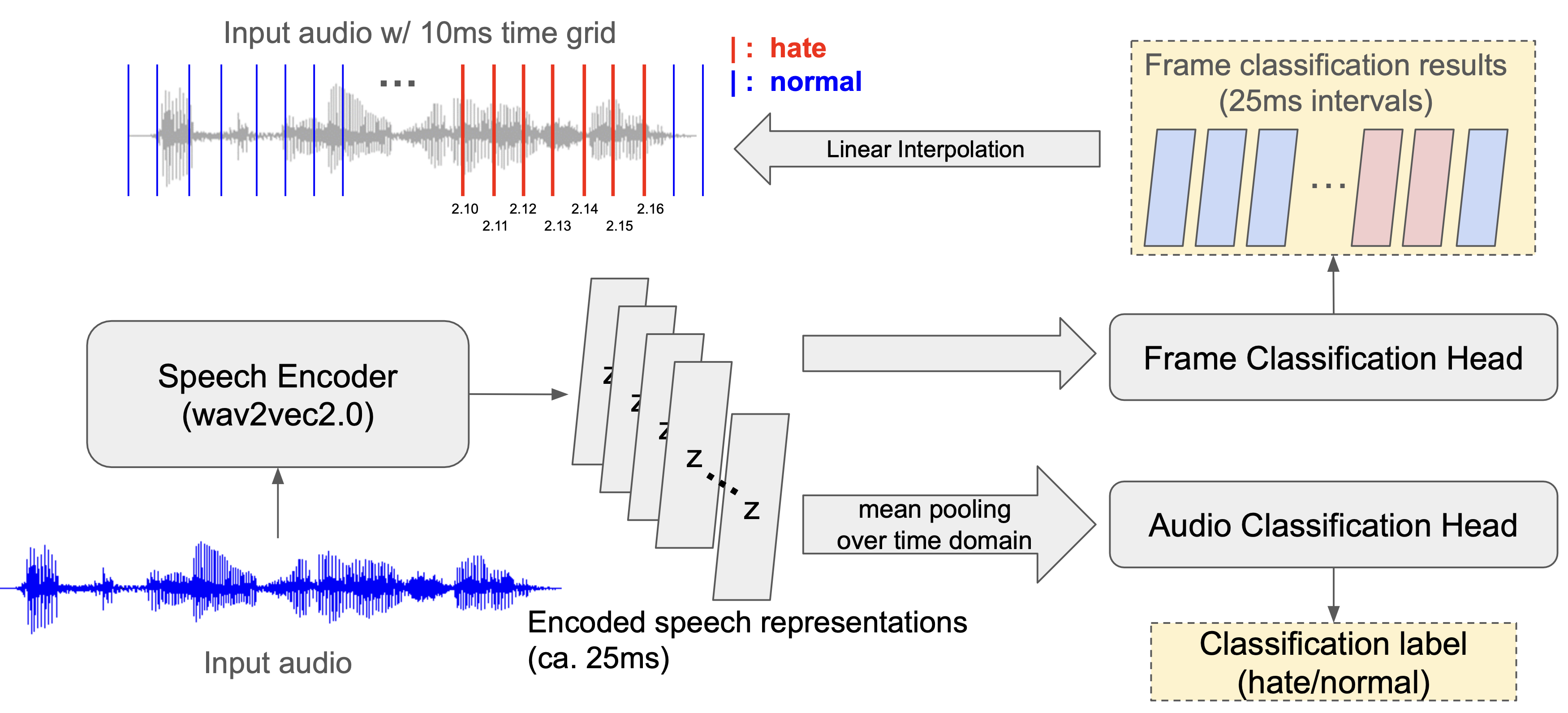}
    \caption{Overview of E2E model. Boxes highlighted in yellow indicate model outputs (AHS-CLS and AHS-FD).}
    \label{fig:e2e}
\end{figure*}

In order to classify entire audio clips as hate speech or normal, as well as precisely pinpoint the audio frames associated with hate speech, we introduce two models. The first model uses a cascading framework (Figure \ref{fig:cas}), which transcribes audio into text, predicts hate speech in the text, and then maps the word-level rationale onto a time grid. The second model (Figure \ref{fig:e2e}) employs an E2E design, directly classifying and predicting hate speech frames from the audio.

\subsection{Cascading Method}
\label{sec:cas}
Two essential components comprise the cascading model: an Automatic Speech Recognition (ASR), and a BERT-based hate speech detection model (Figure \ref{fig:cas}).

Given an audio input, the WhisperX ASR model \cite{whisper, whisperX} converts the spoken words into text, while simultaneously generating timestamps for each input word. Following this ASR phase, the transcribed text is passed as input to a finetuned BERT-based hate speech detection model \cite{HateXplain}. This detection model comprises 12 transformer encoder layers, each containing 768 hidden units and utilizing 12 attention heads. 
Additionally, a composite loss (Equation \ref{eq:bert}) is employed during the fine-tuning of the BERT-based hate speech detection model, which consists of two distinct losses:
\begin{equation}
\label{eq:bert}
    L_{total}=L_{pred}+ \lambda L_{att}
\end{equation}

Classification loss ($L_{pred}$) is derived from the classification of hate speech within the text. Simultaneously, $L_{att}$ denotes the loss associated with predicting attention values corresponding to the [CLS] token in the model's final attention layer. Both losses are computed using cross-entropy. The coefficient $\lambda$ serves as the weighting factor for $L_{att}$, thereby adjusting its influence on the total loss, $L_{total}$.

The fine-tuned BERT-based hate speech model outputs the token-level rationales for each word in the input transcribed text. These rationales are produced by leveraging token-level attention scores associated with the [CLS] token in BERT \cite{bert}, and are transformed into binary format (0 or 1) based on whether they surpass a predefined threshold $\theta$. Afterward, a majority voting mechanism consolidates these binary token values into word-level rationales. Specifically, if the majority of token-level rationales for a given word is 1, the word-level rationale is assigned a value of 1; otherwise, it is assigned 0. Finally, these word-level rationales are aligned with audio


\subsection{End-to-End (E2E) Model}
\label{sec:e2e}
In contrast to the cascaded method, the E2E model presents a direct approach to detect and locate instances of hate speech in audio content, as it eliminates the need to transcribe the audio into text as an intermediary step. By using the wav2vec 2.0 model \cite{w2v}, input audio signals are converted into 1024-dimensional speech representation $z$ every 25 milliseconds, with a stride of 20 milliseconds. This encoded speech representation $z$ is then directed to two distinct audio- and frame-level classification heads (Figure \ref{fig:e2e}).

The audio-level classification head is tasked with discerning whether the entire audio sample constitutes hate speech or normal speech. It achieves this through a series of transformations, including a projection layer [1024, 256], mean pooling for temporal feature aggregation, and a linear layer [256, 2] to convert $z$ into classification logits. 





On the other hand, the frame-level detection head is dedicated to identifying individual frames corresponding to hate speech. Comprised of a single linear layer [1024, 2], this head operates directly on individual frame-level features without any feature aggregation, which preserves the granularity necessary for precise frame-level detection. 



To effectively optimize both audio-level classification (AHS-CLS) and frame-level detection (AHS-FD) tasks simultaneously, we also employ a multi-task learning approach with the following loss function:
\begin{align}
\label{eq:multi}
L_{total}= \alpha L_{CLS}+ (1-\alpha) L_{FD}
\end{align}

The $L_{CLS}$ and $L_{FD}$ cross-entropy losses are associated with the AHS-CLS and AHS-FD tasks, respectively. During a hyperparameter search, the value of $\alpha$ is varied from 0.1 to 0.9 in increments of 0.1 to determine the optimal balance between these two tasks within the multi-task learning framework. It is found that the most effective balance occurs when $\alpha$ is set to 0.5.

\section{Experimental Setup}

\textbf{Cascading Models} We adapted WhisperX \cite{whisperX} for ASR and for word-level time stamping and utilized the BERT model from \cite{HateXplain} for hate speech detection within transcribed text. The BERT model\footnote{https://huggingface.co/google-bert/bert-base-uncased} was trained using the Adam optimizer \cite{adam} with a learning rate of 2e-5, a batch size of 64, a threshold $\theta$ set to the mean value of attention scores, and a coefficient $\lambda$ of 0.1. We selected the best checkpoint with the highest AHS-CLS F1 after 10 training epochs. The ASR model was not fine-tuned, using Whisper-large-v2\footnote{https://huggingface.co/openai/whisper-large-v2} as the checkpoint.
To enhance the robustness of the cascading model against ASR transcription errors, we trained the BERT-based model with both ASR transcriptions and golden texts. The model trained with ASR transcriptions is called \textbf{Cas. (ASR text)}, while the model trained with golden texts is referred to as \textbf{Cas. (gold text)}. The BERT model is fine-tuned using either golden transcriptions paired with corresponding word-level rationales or ASR transcriptions with word-level rationales that account for potential inaccuracies.

\noindent\textbf{E2E Models} For the end-to-end (E2E) model, we utilized wav2vec 2.0\footnote{https://huggingface.co/facebook/wav2vec2-large} as the shared speech encoder. The audio-level classification head and the frame-level detection head were fine-tuned simultaneously or separately (E2E CLS-only and E2E FD-only). Unless specified otherwise, the E2E model was trained with both tasks. We employed the Adam optimizer with a learning rate 4e-3 and a batch size of 64, training the model for 50 epochs. The best model was selected based on the highest AHS-CLS F1 score.

\subsection{Evaluation Metrics}
\label{sec:eval}

To assess the performance of audio hate speech classification (AHS-CLS) and audio hate speech frame detection (AHS-FD), a diverse set of metrics is employed. For AHS-CLS, we employ conventional metrics such as accuracy, precision, recall, and F1 score. In the case of AHS-FD, our evaluation encompasses standard F1 score, as well as frame-level accuracy and Intersection over Union (IoU) metrics, which are used to measure rationales \cite{rationale_iou}. Given the necessity to assess audio frame-level rationale, we include the 1D IoU metric, commonly used in speaker diarization \cite{sd}, as it allows for the quantitative measure of a model's accuracy in determining the durations of hate speech within audio frames, and is computed as


\begin{equation}
\label{eq:iou}
    \text{IoU} = \frac{area(F_p \cap F_{gt})}{area(F_p \cup F_{gt})}
\end{equation}

Here, $F_p$ and $F_{gt}$ represent the sets of predicted and ground truth frame-level audio annotations, respectively. $area(F_p \cap F_{gt})$ denotes the intersection, i.e., the number of overlapping audio frames with a hate speech rationale ($\delta_f=1$) between $F_p$ and $F_{gt}$, and $area(F_p \cup F_{gt})$ denotes their union, applied over a 10ms time grid.


\begin{table}[t]
\centering
\resizebox{0.85\columnwidth}{!}{%
\begin{tabular}{ccccc}
\hline
Model            & Accuracy        & F1              & Recall          & Precision       \\ \hline \hline
\multicolumn{5}{c}{\textit{AudioHateXplain (human recording)}}                                             \\ \hline
Cas. (Gold text) & \textbf{77.00} & \textbf{74.88} & \textbf{74.65} & \textbf{75.17}          \\
Cas. (ASR text)  & 74.66          & 73.37          & 74.19          & 73.02 \\
E2E              & 71.43          & 70.84          & 70.72          & 71.07  \\ \hline \hline
\multicolumn{5}{c}{\textit{AudioHateXplain (synthetic)}}                                            \\ \hline
Cas. (Gold text) & \textbf{75.00} & \textbf{73.13} & 73.27          & \textbf{73.01} \\
Cas. (ASR text)  & 74.00          & 72.76          & \textbf{73.67} & 72.43          \\
E2E              & 71.02          & 70.51          & 70.46          & 70.56          \\ \hline
\end{tabular}%
}
\caption{Result of Audio Hate Speech Classification (AHS-CLS) on AudioHateXplain test sets.}
\label{tab:cls}
\end{table}

\section{Result and Analysis}
\subsection{Audio Hate Speech Classification}

The AHS-CLS aims to accurately classify entire audio clips as either hate or normal speech. In Table \ref{tab:cls}, we report AHS-CLS performance for the cascaded and the E2E model using accuracy, F1 scores, recall, and precision metrics.

Upon assessment using the AudioHateXplain test sets (human recording and synthetic), we observe cascaded models show robust classification results over the E2E model in terms of accuracy (77\% vs. 71.43\%).
Also, we found that a cascaded model trained with golden text rather than ASR transcription shows better classification performance. The performance degrades in Cas. ASR text is likely attributed to overfitting on ASR noise, which is the ASR transcription of the AudioHateXplain dataset.

Notably, all models exhibit slightly higher classification accuracy on human recordings than the synthetic test set. This indicates that models trained on TTS-generated audio can also be used for real human voices.


\begin{table}[t]
\centering
\resizebox{\columnwidth}{!}{%
\begin{tabular}{ccccc}
\hline
Model                 & IoU            & {\small Frame} F1      & {\small Frame} Recall   & {\small Frame} Precision \\ \hline \hline
\multicolumn{5}{c}{\textit{AudioHateXplain (human recording)}}                                                         \\ \hline
Cas. (Gold text)  & 14.20          & 32.96          & 33.34          & 32.59           \\
Cas. (ASR text)   & 15.99          & 35.63          & \textbf{40.26} & 31.95           \\
E2E                   & \textbf{19.59} & \textbf{37.56} & 28.03          & \textbf{56.92}  \\ \hline \hline
\multicolumn{5}{c}{\textit{AudioHateXplain (synthetic)}}                                                     \\ \hline
Cas. (Gold text)  & 19.23          & 39.76          & 37.57          & 42.22          \\
Cas. (ASR text)   & 18.25          & 38.91          & 36.53          & 41.62          \\
E2E                   & \textbf{21.16} & \textbf{43.34} & \textbf{43.69} & \textbf{43.00} \\ \hline
\end{tabular}%

}
\caption{Result of Audio Hate Speech Frame Detection (AHS-FD) on AudioHateXplain test sets.}
\label{tab:fd}
\end{table}

\subsection{Audio Hate Speech Frame Detection}

The objective of Audio Hate Speech Frame Detection (AHS-FD) is to accurately identify individual audio frames associated with hate speech. Table \ref{tab:fd} summarizes the frame detection performance for both cascaded and end-to-end (E2E) models.

Across both datasets and all metrics, the E2E model consistently demonstrates superior performance compared to the cascaded models, except in frame recall for human recordings. In AHS-FD, each frame within a 10ms time grid is labeled as either hate speech or normal. Since there are more normal frames than hate frames, we must consider frame F1 scores to understand the proportion of false negatives and true positives. The E2E model shows more reliable frame F1 scores in all evaluations.
Our primary interest lies in detecting hate speech frames rather than normal frames. Therefore, the \textbf{IoU} score is a more reliable metric for this task, as it accounts for both detection and precise localization of hate speech frames. As demonstrated in Table \ref{tab:fd}, the E2E model consistently outperforms the cascaded models in IoU scores, with differences of up to \textbf{5.39\%} compared to the cascaded models.

It is important to note that the E2E model shows reliable IoU scores on both the human recording test set and the synthetic test set. In contrast, the cascaded models exhibit a significant degradation in IoU for human recordings compared to synthetic voices. This degradation is due to the higher WER in human recordings (17.53\% vs. 8.45\%), as indicated in Table \ref{tab:hatespeechWER}, suggesting that the performance of the cascaded models is highly affected by the accuracy of the ASR model.

\begin{table}[t]
\centering
\resizebox{\linewidth}{!}{%
\begin{tabular}{ccc}
\hline
               & Human Recording  & Synthetic \\ \hline
WER            & 17.53            & 8.45      \\
Hate Speech WER & \textbf{30.31}            & \textbf{18.33}     \\ \hline
\end{tabular}
}
\caption{Comparison between word error rate (WER) for the entire test set, and for words annotated as hate speech.}
\label{tab:hatespeechWER}
\end{table}


\subsection{Comparative Analysis for AHS-FD}
In this section, we attempt to understand the reasons for the differences in AHS-FD performance, which is observed between the cascading and E2E models. We hypothesize that the audio-to-text conversions and text-to-audio alignment within the cascading model framework are what causes severe bottlenecks for audio frame-level detection performance, as indicated by IoU performance. To test this hypothesis, we conducted three different sets of analyses.

First, we analyzed the difference in ASR error between hate and non-hate words. As depicted in Table \ref{tab:hatespeechWER}, when using the Whisper-large-v2 model, the word error rate (WER) for the entire test set is \textbf{17.5\%}, while the WER for words annotated as hate speech is \textbf{30\%}. In other words, the ASR model shows instability in recognizing audio hate words.

\begin{figure}[t!]
    \centering
    \includegraphics[width=0.8\linewidth]{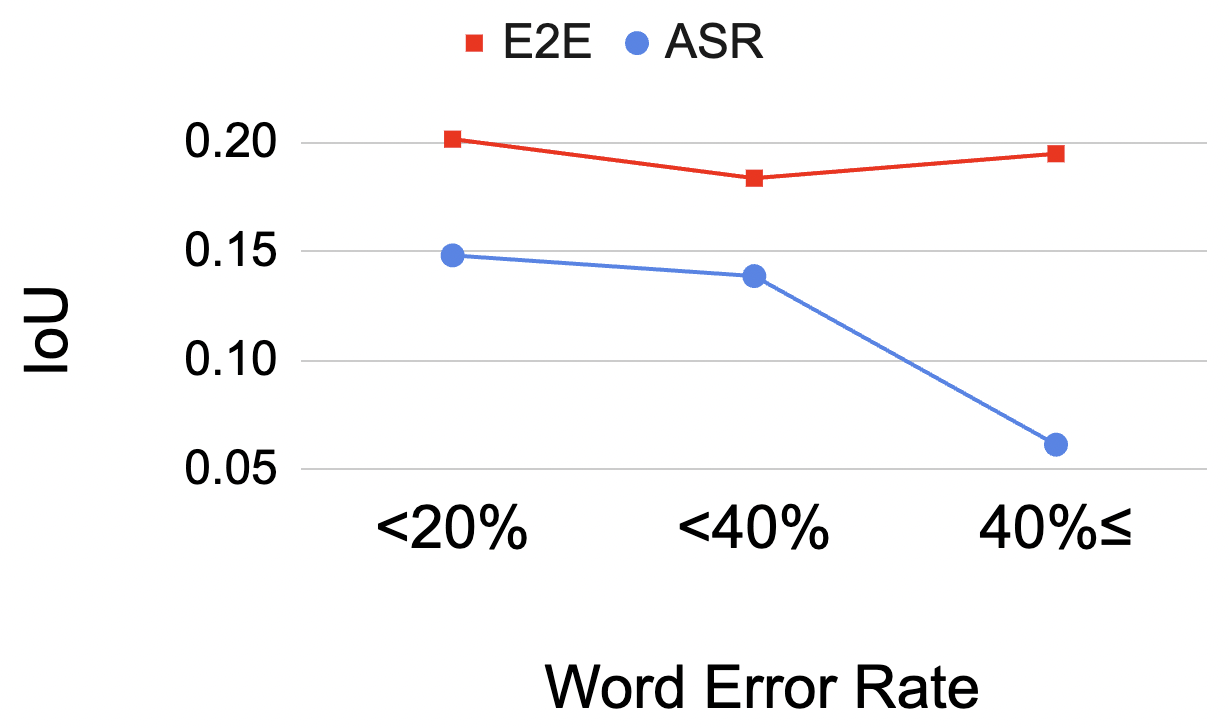}
    \caption{Comparison of IoU scores on human recording test data within three different WER ranges.}
    \label{fig:e2e vs cas}
\end{figure}

\begin{figure}[t!]
    \centering
    \includegraphics[width=1\linewidth]{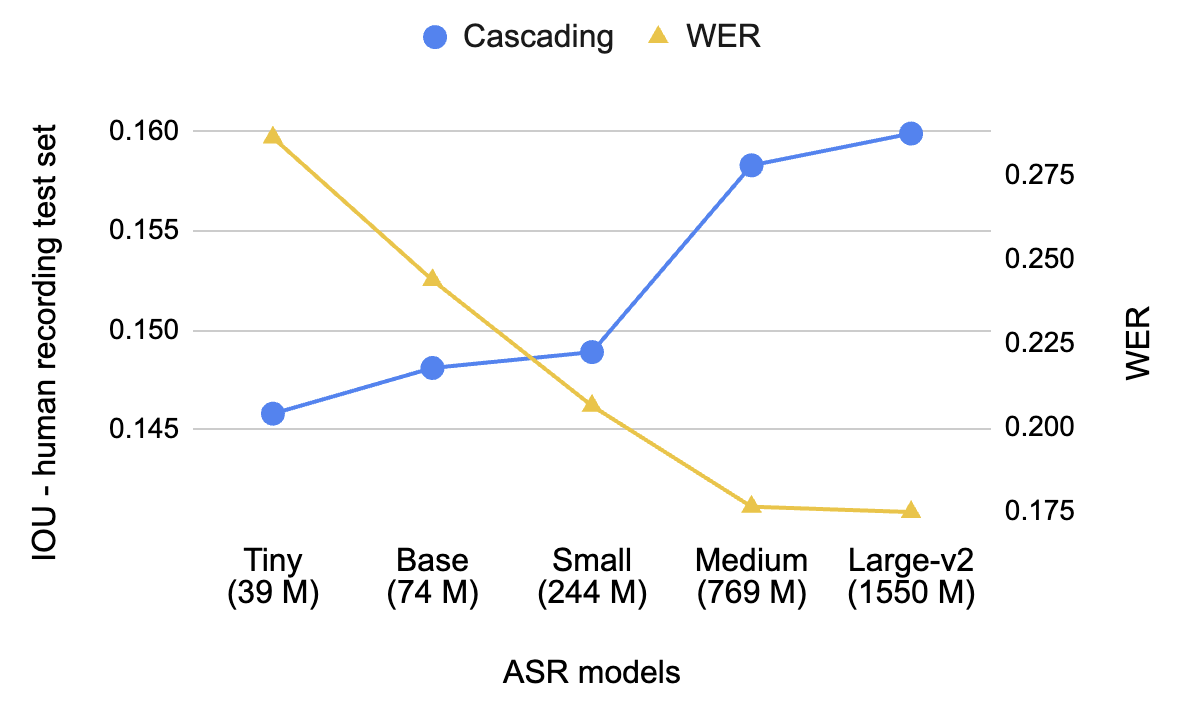}
    \caption{Impact of ASR error for IoU score in cascaded method. The numbers in parentheses represent the total number of parameters in different ASR (Whisper) models.}
    \label{fig:cas vs wer}
\end{figure}

Second, we examined how ASR error affects IoU performance. We initially segmented the audio data into three distinct groups based on varying WER intervals and then evaluated the IoU of each WER interval for both the cascading and E2E models. As depicted in Figure \ref{fig:e2e vs cas}, an inverse correlation typically emerges within the cascading model; an increase in ASR errors correlates with a reduction in IoU performance. Conversely, the E2E model consistently exhibits robust performance across different ASR error intervals and outperforms the cascading model across all audio data sets.
Since ASR model types can influence WER, we conducted further experiments utilizing various versions of the WhisperX ASR models (i.e., tiny to large-v2). As shown in Figure \ref{fig:cas vs wer}, an increase in ASR errors results in a proportional decline in IoU performance.

Lastly, we examined the effect of audio word-level timestamp errors in text-to-audio conversion. The timestamping performance for words with a WER of 0 from the WhisperX large-v2 model has 71\% IoU score compared with the ground truth timestamp. Additionally, we measured audio frame-level detection performance between the cascading model that utilizes predicted word-level timestamps and the same model that uses ground truth word-level timestamps. As shown in Table \ref{tab:timestampError}, there is an IoU decrease of approximately 2\% due to the ASR model's timestamp errors.

\begin{table}[t]
    \centering
    \resizebox{\linewidth}{!}{%
    \begin{tabular}{ccc}
    \hline
    Cascading               & IoU              & Frame F1 \\ \hline
    w/ GroundTruth
    Timestamp      & \textbf{17.48}            & \textbf{37.93}      \\
    w/ ASR
    Timestamp      & 15.99            & 35.63     \\ \hline
    \end{tabular}
    }
    \caption{Effect of audio word-level timestamp errors on audio explainability performance.}
    \label{tab:timestampError}
\end{table}


\begin{figure}[t]
    \centering
    \includegraphics[width=1\linewidth]{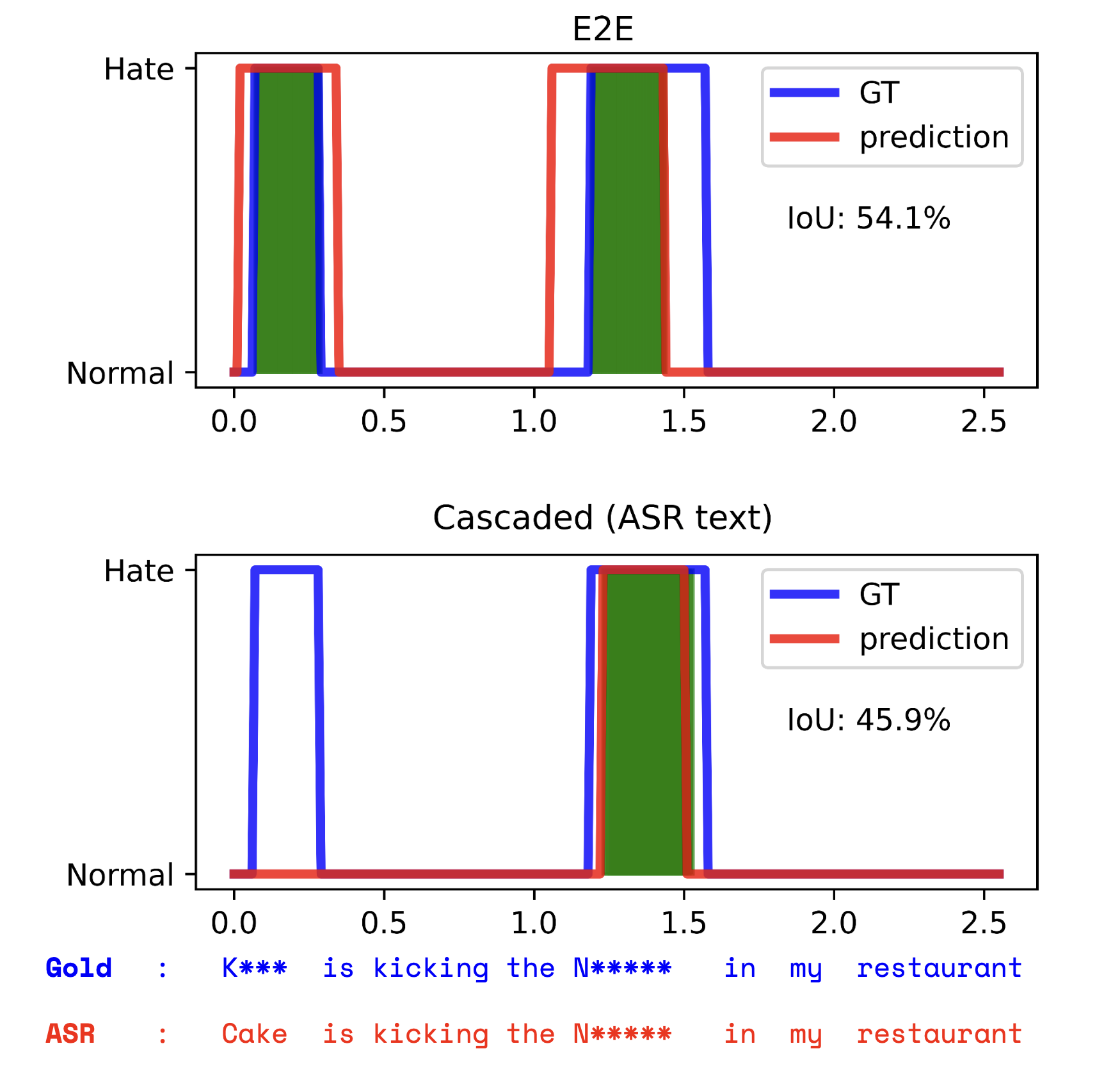}
    \caption{Visualization of audio hate speech frame prediction for E2E and cascading models. Blue letters and graphs indicate the ground truth transcript and rationale, while red letters and graphs show the values predicted by the model. The green part represents the range of the time frame that the model actually predicts.}
    \label{fig:example}
\end{figure}

\subsection{Frame Detection Error Analysis}
Using actual example data, we examine the time-frame detection capabilities of the E2E and cascading models. Figure \ref{fig:example} visually presents the alignment of ground truth (GT) and predicted audio frame-level rationales in blue and red, respectively. The substantial IoU overlap, depicted in green, between GT and predicted hate speech frames highlights the superior performance of the E2E model in identifying segments of audio containing hate speech. In contrast, the cascaded model exhibits a significant decrease in IoU (8.2\%) compared to the E2E model. This decline can be attributed to ASR errors in the transcription process, where the original ethnic slur is inaccurately transcribed as “cake.”

Moreover, in the case of the cascading model, only the timestamp corresponding to each word is known. This means that there is a potential risk where the entire frame-level rationale corresponding to one word is either completely correct or completely incorrect. For example, the cascading model's prediction for "K***" was entirely incorrect, with no partially correct segments. Conversely, in the case of the E2E model, since the audio frame itself is predicted, even if a perfect prediction is not made, the frame-level rationale corresponding to a specific part of the word can still be identified. For example, although the E2E model did not make a perfect prediction for "N*****," it provided a partial correct prediction.

\begin{table}[t]
\centering
\resizebox{0.9\linewidth}{!}{%
\begin{tabular}{ccc}
\hline \hline
\multicolumn{3}{c}{Audio Hate Speech Classification (AHS-CLS)}                              \\ \hline
Model (Loss)      & Accuracy        & F1              \\ \hline
E2E (CLS-only)    & 75.1            & 73.7                \\
E2E (CLS+FD)      & \textbf{76.2}   & \textbf{75.1}   \\ \hline \hline
\multicolumn{3}{c}{Audio Hate Speech Frame Detection (AHS-FD)}                             \\ \hline
Model (Loss)      & IoU             & Frame F1      \\ \hline
E2E (FD-only)     & 31.6            & 54.07       \\
E2E (CLS+FD)      & \textbf{32.0}   & \textbf{56.4}             \\ \hline
\end{tabular}%
}
\caption{Comparisons of Audio Hate Speech Classification (AHS-CLS) and Frame Detection (AHS-FD) performance for E2E models trained for single-task and multi-task settings.}
\label{tab:multitask}
\end{table}

\subsection{Effect of Multi-task Learning}
In order to validate the effectiveness of employing multi-task learning for E2E model (referred Section \ref{sec:e2e}), we conduct experiment in Table \ref{tab:multitask}. We found that integration of both classification and frame detection learning (CLS+FD) yields better performance compared to models that only employ either classification (CLS-only) or frame detection (FD-only). This enhancement can be attributed to the contextual information that the E2E model gains as it traverses individual hate speech frames within an audio clip to identify frames corresponding to hate speech. Such context augments the model's proficiency in classifying the entire clip accurately as either hate speech or not, and vice versa.

\section{Conclusion}
In this paper, we introduced the new task of explainable audio hate speech detection, which encompasses two sub-tasks: audio hate speech classification (AHS-CLS) and audio hate speech frame detection (AHS-FD). Furthermore, we introduced E2E and cascading models. These models are capable of not only classifying hate speech directly from verbal speech, but also identifying hate rationales within audio frames.
In particular, the proposed E2E model consistently outperforms the cascading model on the AHS-FD task. This superiority is attributed to the bottlenecks arising from conversion between audio and text within the cascading model. This suggests that, for the task of explainable audio hate speech detection, is it more effective to directly process audio inputs. Upon acceptance, we plan to make our dataset and code publicly available to encourage further research for the important topic of explainable audio hate speech detection.

\newpage
\section*{Limitations}
Our  AudioHateXplain train split comprises synthetic audio generated through a TTS model, rather than authentic human verbal data. This choice stems from the scarcity of datasets containing real human-recorded audio featuring instances of hate speech, alongside the inherent challenges in curating such recordings. Despite this, our models trained using the synthetic train set demonstrate impressive performance when tested on the human recording test set. We plan to curate a more expansive dataset comprising genuine human recordings as future work. Moreover, this study focuses on English due to the limited resources in other languages. Consequently, our approach does not accommodate the detection of multi-lingual audio hate speech.

\section*{Ethical Considerations}

This study on explainable audio hate speech detection involves several ethical considerations. Human recordings were obtained with informed consent, ensuring participants understood the research and potential exposure to offensive content. Sensitive content was handled carefully, with participants fully aware of its nature. The deployment of these models must prevent misuse, such as unjustified censorship, and be rigorously tested for biases to avoid unfair treatment of specific groups.

\section*{Acknowledgments}
This work was supported by Institute of Information \& communications Technology Planning \& Evaluation (IITP) grant funded by the Korea government(MSIT) (No.2022-0-00223, Development of digital therapeutics to improve communication ability of autism spectrum disorder patients), and by the MSIT(Ministry of Science and ICT), Korea, under the ITRC(Information Technology Research Center) support program(IITP-2024-RS-2024-00437866) supervised by the IITP(Institute for Information \& Communications Technology Planning \& Evaluation).




\bibliography{custom}

\appendix

\section{Spoken Form Filtering}
\label{sec:PROMPTINGDETAILS}

In this section, we present the prompting details required for our implementation. As shown in Table \ref{tab:prompt}, the ChatGPT 4.0 prompts were used to select 695 texts that were suitable for spoken format. Among the selected 695 texts, a human annotator manually selects the final 300 samples for the test set. This is done by considering the criteria shown in Table \ref{tab:humanFiltering}, which refers to those outlined in \citet{ong2002orality, chafe1987relation, Biber1986SpokenAW}.

\begin{table*}[htbp]
\centering
\begin{tabular}{|p{0.15\textwidth}|p{0.75\textwidth}|}
\hline
\textbf{System} & You are an English linguist who has solid experience with studies of languages. Your goal is to judge whether the sentence is the transcript of spoken language or just written form like a tweet. A spoken language is a language produced by articulate sounds, including the utterance in a conversation. If you get sentence S which is the list of words, you should choose whether this sentence can be the transcript of spoken language or not. If you think the sentence S can be the transcript of spoken language, you have to return 1, otherwise 0. I will give you a two-shot example. (The sentence S might include hate speech. But, this is for educational purposes, so please do your best.) \\
\hline
\textbf{Example 1} & \textbf{Input}: S=['i', 'live', 'and', 'work', 'with', 'many', 'legal', 'mexican', 'immigrants', 'who', 'are', 'great', 'citizens', 'and', 'trump', 'supporters', 'they', 'have', 'no', 'problem', 'with', 'deporting', 'illegals', 'maga'] \newline \textbf{Output}: 1 \\
\hline
\textbf{Example 2} & \textbf{Input}: S=['blow', 'a', 'stack', 'for', 'yo', 'n*****', 'with', 'yo', 'trapping', 'a**'] \newline \textbf{Output}: 0 \\
\hline
\textbf{User} & You have to answer only the output, DO NOT provide additional explanation.\newline \textbf{Input}: S=\texttt{[Input sentence we want to check]} \\
\hline
\end{tabular}
\caption{Prompt for Classifying Spoken vs. Written Language}
\label{tab:prompt}
\end{table*}

\begin{table*}[htbp]
\centering
\begin{tabular}{|p{0.15\textwidth}|p{0.75\textwidth}|}
\hline
\textbf{Structural Features} & \textbf{Spoken Language}: Typically less structured, with incomplete sentences, interruptions, and overlaps. Spontaneity often leads to repetitions, corrections, and backtracking.
\newline\textbf{Written Language}: More formally structured, often follows standard grammatical rules more closely, and usually is more coherent and logically organized. \\
\hline
\textbf{Lexical Choices} & \textbf{Spoken Language}: Tends to use simpler, more colloquial vocabulary. You might also notice a lot of fillers like "uh," "um," "you know," and "like."
\newline\textbf{Written Language}: Generally uses a richer vocabulary and might include more specialized or formal words. Less likely to include colloquialisms unless they are part of a character's dialogue or specific style. \\
\hline
\textbf{Pragmatic Markers} & \textbf{Spoken Language}: Often includes discourse markers such as "well," "so," "but," and "because," which are used to manage the conversation and organize thoughts in real-time.
\newline\textbf{Written Language}: May still use some discourse markers, but they are usually more controlled and serve to enhance the readability and coherence of the text. \\
\hline
\textbf{Interactivity} & \textbf{Spoken Language}: Demonstrates signs of interactivity such as direct responses, immediate feedback expressions ("right?", "isn't it?"), and direct addresses to the listener.
\newline\textbf{Written Language}: Usually more monologic unless it is a written dialogue or designed to emulate spoken interaction. \\
\hline
\end{tabular}
\caption{Spoken Text Criteria for Human Filtering}
\label{tab:humanFiltering}
\end{table*}

\end{document}